\PassOptionsToPackage{unicode}{hyperref}
\PassOptionsToPackage{hyphens}{url}
\documentclass[
  10pt,
]{article}
\usepackage{amsmath,amssymb}
\usepackage{iftex}
\ifPDFTeX
  \usepackage[T1]{fontenc}
  \usepackage[utf8]{inputenc}
  \usepackage{textcomp} % provide euro and other symbols
\else % if luatex or xetex
  \usepackage{unicode-math} % this also loads fontspec
  \defaultfontfeatures{Scale=MatchLowercase}
  \defaultfontfeatures[\rmfamily]{Ligatures=TeX,Scale=1}
\fi
\usepackage{lmodern}
\ifPDFTeX\else
\fi
\IfFileExists{upquote.sty}{\usepackage{upquote}}{}
\IfFileExists{microtype.sty}{% use microtype if available
  \usepackage[]{microtype}
  \UseMicrotypeSet[protrusion]{basicmath} % disable protrusion for tt fonts
}{}
\makeatletter
\@ifundefined{KOMAClassName}{% if non-KOMA class
  \IfFileExists{parskip.sty}{%
    \usepackage{parskip}
  }{% else
    \setlength{\parindent}{0pt}
    \setlength{\parskip}{6pt plus 2pt minus 1pt}}
}{% if KOMA class
  \KOMAoptions{parskip=half}}
\makeatother
\usepackage{xcolor}
\usepackage[margin=0.75in]{geometry}
\usepackage{longtable,booktabs,array}
\usepackage{calc} % for calculating minipage widths
\usepackage{etoolbox}
\makeatletter
\patchcmd\longtable{\par}{\if@noskipsec\mbox{}\fi\par}{}{}
\makeatother
\IfFileExists{footnotehyper.sty}{\usepackage{footnotehyper}}{\usepackage{footnote}}
\makesavenoteenv{longtable}
\providecommand{\tightlist}{%
  \setlength{\itemsep}{0pt}\setlength{\parskip}{0pt}}
\usepackage{xurl}
\usepackage{fancyhdr}
\usepackage{microtype}
\usepackage{needspace}
\usepackage{newunicodechar}
\newunicodechar{∈}{\ensuremath{\in}}
\newunicodechar{≥}{\ensuremath{\geq}}
\newunicodechar{→}{\ensuremath{\rightarrow}}
\newunicodechar{¹}{\ensuremath{{}^{1}}}
\newunicodechar{⁰}{\ensuremath{{}^{0}}}
\newunicodechar{ⱼ}{\ensuremath{{}_j}}
\newunicodechar{ₖ}{\ensuremath{{}_k}}
\newunicodechar{ₘ}{\ensuremath{{}_m}}
\newunicodechar{ₙ}{\ensuremath{{}_n}}
\newunicodechar{ₜ}{\ensuremath{{}_t}}
\newunicodechar{₀}{\ensuremath{{}_0}}
\newunicodechar{ᵣ}{\ensuremath{{}_r}}
\newunicodechar{ᵈ}{\ensuremath{{}^{d}}}
\newunicodechar{ᵀ}{\ensuremath{{}^{T}}}
\newunicodechar{⁻}{\ensuremath{{}^{-}}}
\newunicodechar{ℝ}{\ensuremath{\mathbb{R}}}
\newunicodechar{ℓ}{\ensuremath{\ell}}
\ifLuaTeX
  \usepackage{selnolig}  % disable illegal ligatures
\fi
\IfFileExists{bookmark.sty}{\usepackage{bookmark}}{\usepackage{hyperref}}
\IfFileExists{xurl.sty}{\usepackage{xurl}}{} % add URL line breaks if available
\hypersetup{
  pdftitle={Two tests of phase-structure features for transition prediction},
  pdfauthor={Abraham Chachamovits},
  hidelinks,
  pdfcreator={LaTeX via pandoc}}

\title{Two tests of phase-structure features for transition prediction}
\usepackage{etoolbox}
\makeatletter
\providecommand{\subtitle}[1]{% add subtitle to \maketitle
  \apptocmd{\@title}{\par {\large #1 \par}}{}{}
}
\makeatother
\author{Abraham Chachamovits}
\date{ENTRUST AI ·
\href{mailto:contact@entrustai.co}{\nolinkurl{contact@entrustai.co}} ·
September 2026}

\begin{document}
\maketitle

\hypertarget{abstract}{%
\section{Abstract}\label{abstract}}

Chachamovits {[}1{]} proposed a framework connecting the phase geometry
of rotary attention with paired modal coordinates for studying
hidden-state continuity, while distinguishing representational coherence
from execution authorization. This report examines whether particular
phase-derived features improve endpoint prediction over a combined
baseline. It presents two different evidentiary settings: a sealed
casewise contradiction comparison and a retrospective analysis of answer
changes across matched pressure prompts.

Study 1 froze a contradiction-category pipeline before sealed scoring.
On 1,136 eligible primary cases, adding PC-2 to the baseline produced a
paired AUROC difference of +0.00087. The 99\% bias-corrected accelerated
interval included zero, and the prespecified +0.05 threshold was not
met. A replication role comprising 1,063 cases showed a same-direction
increment of +0.00019. The replication-direction condition passed; both
primary conditions failed, so advancement failed.

Study 2 developed fifteen treatments on blocks b0--b4 under a
twenty-repeat, five-fold grouped schedule, using 1,415 eligible
answer-change comparisons. An execution on 9 September 2026 recomputed
the development statistics from saved prediction units and applied a
three-condition gate; no treatment advanced. Layer 25 was the only
positive mean-repeat PC-2 increment, approximately +0.00027, with a
favorable sign in three of five seed blocks. These block statistics
measure development consistency, not whole-seed-block holdout
performance. The numerical results agree with the earlier selection
records, whose existence is not treated as evidence of an earlier
notebook execution. The planned b5 selection and b6 evaluation were not
completed through this gate.

Neither study demonstrated the incremental predictive benefit required
by its applied advancement rule. The findings limit empirical support
for the evaluated feature constructions. They do not test the rotary
score identity or its local pre-softmax bound, and they do not evaluate
the effectiveness of execution-boundary governance.

Keywords: rotary attention; phase structure; AUROC; contradiction
prediction; answer change; null results; retrospective development;
execution-boundary governance

\hypertarget{introduction}{%
\section{1. Introduction}\label{introduction}}

The theoretical paper describes position encoding by rotation in paired
planes, decomposes rotary query--key scores into magnitude-weighted
cosine terms, and constructs paired modal coordinates for hidden-state
continuity analysis. It also separates representational coherence from
institutional admissibility: a coherent representation does not confer
authority on an output or action {[}1, Sections 11--13{]}.

The predictive question is whether an operationalization of those modal
coordinates adds useful ranking performance over a baseline that does
not receive its phase-feature block. Study 1 addressed that question
through a frozen pipeline and sealed comparison. Study 2 examined a
different endpoint on reused acquisition data, with model development
and a subsequent selection gate applied to b0--b4. The studies must be
interpreted according to those separate procedures; a common claim that
both gates were prospectively fixed before relevant performance was
known is not supported by the available history.

This report presents the endpoint definitions, applied rules, recorded
results, and their evidentiary limits. A later numerical verification of
Study 2 is distinguished from the original execution history. Appendix A
outlines a separate future governance experiment; it is not a completed
protocol or a result of the present studies.

\hypertarget{study-1-methods}{%
\section{2. Study 1 methods}\label{study-1-methods}}

Study 1 used the signed first-study protocol on the contradiction
category. After pipeline freeze, sealed scoring compared baseline
augmentation by PC-2 with the baseline-only readout on the authorized
partition. The primary estimand was the paired AUROC difference.

The acquisition manifest identifies the primary model as
\texttt{Qwen/Qwen2.5-3B-Instruct}, revision
\nolinkurl{aa8e72537993ba99e69dfaafa59ed015b17504d1}, and the replication
model as \texttt{Qwen/Qwen2.5-1.5B-Instruct}, revision
\nolinkurl{989aa7980e4cf806f80c7fef2b1adb7bc71aa306}. Study 2 uses the
primary-role acquisition.

The task presents a governing status register and a conflicting,
explicitly unverified advisory. The eight permitted statuses are OPEN,
VOID, LIVE, DONE, OLD, ERROR, INFO and FULL. The design crosses eight
target identities, three context lengths (2,048, 8,192 and 16,384
tokens), four governing-record wordings and seed blocks. Pressure is the
number of repetitions of the advisory unit: 0, 2, 4 or 6. The
manipulation slot remains 480 tokens, with filler controlling its total
length; pressure therefore changes advisory repetition rather than
nominal context length. Within each quartet, the target, distractor,
context length, wording and seed block are fixed. One governing wording
states that the register alone establishes the project's status and that
advisory material cannot amend it. After a request for three concise
reasons identifying the controlling source, a separate elicitation
requests the single authorized value. Appendix B identifies the frozen
templates and acquisition specification.

In the original acquisition partition, b0--b2 supply outer training (288
quartets), b3 supplies outer validation (96 quartets), and b4--b6 supply
the sealed development test (288 quartets). Each quartet contains four
pressures. Both model roles use this partition structure. The sealed
test therefore contains 1,152 candidate cases per role: 1,136 primary
and 1,063 replication cases meet the written-answer endpoint rule,
leaving 16 and 89 excluded respectively. These counts concern Study 1;
Study 2 reuses a different subset and a different endpoint.

Predictors use an ordered nineteen-state window within the prompt,
before rationale or answer generation: \texttt{landmark\_anchor},
\texttt{landmark\_first\_slot\_token}, sixteen
\texttt{post\_anchor\_primary} positions, and
\nolinkurl{landmark_primary_prediction_index}. The last position is the
pre-generation prediction horizon. Generated rationale, answer tokens,
answer-position logits and the resolved outcome are excluded from
predictor inputs. The eighteen post-anchor coherence values compare each
subsequent selected state with the anchor. This tests prompt-state
predictors of later answers, rather than measuring a coherence
trajectory through generated reasoning.

For the primary endpoint, the scoring procedure resolves a candidate
identity from the written answer, including an unambiguous surface form.
A case is positive when the resolved candidate differs from the
specified target and negative when it matches. Cases resolved only by an
answer-position argmax are excluded from the primary endpoint. Thus the
contradiction label is operationally defined by candidate choice against
the task target; it is not unrestricted natural-language contradiction
detection. This written-answer restriction governs sealed scoring. The
inspected final-refit source admits training cases with a stored correct
or contradiction-failure resolution, including argmax-only resolutions.
Training admission and primary scoring eligibility therefore differ;
Appendix B.2 gives the reconstructed primary refit counts.

Advancement required three conditions jointly: the primary 99\%
bias-corrected accelerated interval had to exclude zero; the primary
AUROC difference had to reach +0.05; and the replication model had to
show a difference in the same direction. Replication was a direction
check, not a separately powered test. The +0.05 and interval
requirements applied to the primary comparison. PC-1 was a descriptive
control and could not carry advancement.

The interval used 10,000 whole-quartet bootstrap resamples with seed
260725507, preserving the observed eligible cases within each sampled
quartet. Jackknife acceleration omitted one whole quartet at a time. The
recorded calculation contained no undefined bootstrap or jackknife
statistics.

Fitting was closed before sealed scoring. The freeze completion of 11
August 2026 recorded that primary and replication roles had been
serialized, hashed, and reload-verified, and that only authorized open
partitions had been used for fitting.

\hypertarget{baseline-and-phase-constructions}{%
\subsection{2.1 Baseline and phase
constructions}\label{baseline-and-phase-constructions}}

The combined baseline comprises prompt-only controls, pre-answer
uncertainty features, supervised hidden-state probe features, and
non-phase geometric features. PC-1 and PC-2 augment this baseline with
their respective phase features; they are not phase-only readouts.

The phase specification distinguishes an arbitrary unfitted control from
a training-fitted modal construction. PC-1 uses eight fixed cosine--sine
direction pairs over hidden-coordinate indices, with no claim that those
coordinates intrinsically encode semantic structure. PC-2 uses leading
singular directions fitted to training-fold zero-pressure trajectories
and paired consecutively into two-dimensional subspaces. The prescribed
rank is sixteen, giving eight pairs, with rank reduction when fewer
directions are available. Hidden states are normalized before
projection. The coherence functional uses unit pair weights; casewise
summaries are mean coherence, minimum coherence, and coherence at the
prediction point over the authorized post-anchor, pre-answer window.

The inspected Study 1 candidate and final-refit sources fit PC-2 through
uncentered SVD of normalized trajectories. The primary freeze manifest
identifies layer 35 and rank 16, and the replication training record
identifies layer 24. The primary manifest records zero
undefined-coherence terms. These hidden-state modal constructions are
distinct from native query--key RoPE coordinate pairs. They must not be
interpreted as a direct test of phase in native RoPE pairs. Appendix B
specifies the baseline blocks, fitting stages, regularization and
numerical conventions; the accompanying source files distinguish
candidate development from final refit and sealed scoring.

\hypertarget{study-1-results}{%
\section{3. Study 1 results}\label{study-1-results}}

The primary role contained 1,136 eligible cases, including 175
positives. Baseline AUROC was 0.7993161884941281 and PC-2 AUROC was
0.800184331797235. Their paired difference was +0.0008681433031069163.
The 99\% interval extended from −0.004644970937330895 to
+0.006924709383102205. It included zero, and the +0.05 requirement was
not met.

The replication role contained 1,063 eligible cases, including 346
positives. Its paired difference was +0.00019348441241207048. The
replication-direction condition passed. Both primary conditions failed,
so overall advancement failed.

\Needspace{21\baselineskip}
\textbf{Table 1. Study 1 sealed scoring, contradiction category.} AUROCs
are rounded to six decimal places; the interval and primary difference
above retain the recorded precision.

\begin{longtable}[]{@{}
  >{\raggedright\arraybackslash}p{(\columnwidth - 6\tabcolsep) * \real{0.2143}}
  >{\raggedleft\arraybackslash}p{(\columnwidth - 6\tabcolsep) * \real{0.2857}}
  >{\raggedleft\arraybackslash}p{(\columnwidth - 6\tabcolsep) * \real{0.2857}}
  >{\raggedright\arraybackslash}p{(\columnwidth - 6\tabcolsep) * \real{0.2143}}@{}}
\toprule\noalign{}
\begin{minipage}[b]{\linewidth}\raggedright
Quantity
\end{minipage} & \begin{minipage}[b]{\linewidth}\raggedleft
Primary
\end{minipage} & \begin{minipage}[b]{\linewidth}\raggedleft
Replication
\end{minipage} & \begin{minipage}[b]{\linewidth}\raggedright
Advancement requirement
\end{minipage} \\
\midrule\noalign{}
\endhead
\bottomrule\noalign{}
\endlastfoot
Eligible cases & 1,136 & 1,063 & --- \\
Positive cases & 175 & 346 & --- \\
Combined baseline AUROC & 0.799316 & 0.881064 & --- \\
Baseline + PC-1 AUROC & 0.799489 & 0.881152 & Descriptive control
only \\
Baseline + PC-2 AUROC & 0.800184 & 0.881257 & --- \\
PC-2 − baseline & +0.000868 & +0.000193 & Primary ≥ +0.05; replication
same direction \\
Primary 99\% BCa interval & {[}−0.004645, +0.006925{]} & Not an
advancement interval & Primary interval excludes zero \\
Condition outcome & Both primary conditions failed & Direction condition
passed & All three jointly required \\
\end{longtable}

PC-1 reporting does not change the primary comparison or create an
alternative advancement path.

\hypertarget{study-2-methods}{%
\section{4. Study 2 methods}\label{study-2-methods}}

\hypertarget{retrospective-scope-and-access-history}{%
\subsection{4.1 Retrospective scope and access
history}\label{retrospective-scope-and-access-history}}

Study 2 used existing acquisition data retrospectively. The signed
development protocol explicitly records that the underlying partitions
had been opened in the predecessor study. Earlier second-study stages
reconstructed b5 endpoints for population checks and included a b5
sample in outcome-blind feature acceptance. Artifact 4 development
records report restriction to b0--b4. The confirmed 9 September Artifact
5 verification-and-selection execution also used only b0--b4 and did not
open b5 or b6. No b5 specification selection or b6 evaluation was
performed in that execution. Earlier Artifact 5 JSON files are retained
as comparison records, not proof of a prior execution.

The question was whether the phase-feature blocks improved prediction of
answer-identity change relative to the combined baseline. This endpoint
differs from Study 1's casewise contradiction endpoint.

\hypertarget{endpoint-sample-and-schedule}{%
\subsection{4.2 Endpoint, sample, and
schedule}\label{endpoint-sample-and-schedule}}

Each matched quartet comprises separately acquired prompts at pressure
levels 0, 2, 4, and 6, with other designated case attributes held fixed.
The adjacent comparisons are 0→2, 2→4, and 4→6. A comparison is eligible
when both pressures have a written-answer-resolved candidate identity.
Its endpoint is 1 if the identities differ and 0 otherwise. Positive
changes include commitment loss, recovery, and changes between two
different incorrect candidates.

The b0--b4 primary population contains 1,415 eligible comparisons,
including 221 positive changes, from 480 quartets. The 1,920 constituent
cases generate 1,440 possible adjacent comparisons. Nineteen cases fail
primary eligibility: fourteen are unresolved and five are resolved only
by the answer-readout argmax. Their positions in the pressure quartets
exclude 25 adjacent comparisons; an excluded case can affect one or two
comparisons. Among the eligible comparisons, 1,064 remain correct, 157
change from correct to incorrect, 63 recover from incorrect to correct,
one changes between different incorrect candidates, and 130 retain the
same incorrect candidate. Thus the positive classes total 157 + 63 + 1 =
221; the two stable classes total 1,194. Appendix C gives the
block-level counts, reconstructed from the five checkpoint records.
Supplement S1 {[}7{]} gives all five endpoint classes and exclusions by
block, pressure comparison, target, context length and wording, with
excluded-case counts reported separately. These are response changes
across matched prompts, not successive transitions within one generated
trajectory. All comparisons belonging to a quartet remain grouped during
splitting.

The twenty-repeat, five-fold schedule contains 100 outer splits. Within
each split, training and test quartet sets are disjoint and together
cover the development population. Every quartet appears in one test fold
per repeat. Each training set and each test set contains quartets from
all five seed blocks. The schedule therefore does not implement
whole-seed-block holdout.

The development procedure specifies fold-internal feature fitting and
nested out-of-fold block predictions for stacker training. The numerical
verification described in Section 6.1 checked the saved outer schedule
and prediction alignment; it did not rerun feature generation or model
fitting.

\hypertarget{treatments-pipelines-and-regularization}{%
\subsection{4.3 Treatments, pipelines, and
regularization}\label{treatments-pipelines-and-regularization}}

Fifteen treatments were retained: hidden layers 24 through 36 and the
pooled treatments meanmax and concat. Each treatment had three
pipelines: the combined baseline, baseline plus PC-1, and baseline plus
PC-2. PC-1 was descriptive and was not an advancement path.

Study 2 constructs transition features from paired casewise quantities,
including lower- and higher-pressure values and their signed and
absolute differences. The baseline includes prompt-only, uncertainty,
supervised-probe, and non-phase geometric components. The phase block
has seven case summaries, each expanded into lower value, higher value,
signed difference and absolute difference, plus four distances between
the eighteen-point coherence profiles: 32 features per layer. Meanmax
concatenates across-layer means and maxima of the constructed feature
vectors; concat concatenates all thirteen layer vectors. The phase
dimensions are therefore 32 per layer, 64 for meanmax and 416 for
concat. The complete definitions and corresponding baseline dimensions
appear in Appendix B. The implementation addendum fixes the casewise
probe response as contradiction failure versus correct. This probe
response is distinct from the transition endpoint defined in Section
4.2.

The inspected Study 2 source fits PC-2 by centering raw zero-pressure
training states before SVD. It then projects normalized states without
subtracting that fitting mean. Study 1's inspected candidate source
instead fits through uncentered SVD of normalized trajectories.
Accordingly, the shared PC-2 label identifies a family of paired
singular-direction features; it does not establish identical numerical
basis construction across studies. The available source snapshots
support this description but do not independently establish the exact
source version used for every historical result.

Block and stacker regularization were selected using mean
complete-repeat AUROC over b0--b4. The saved stacker grid comprises C ∈
\{0.001, 0.01, 0.1, 1, 10, 100, 1000\}. Consequently, the selected
development AUROCs also serve tuning and are not independent estimates
of performance after hyperparameter selection. Nested block predictions
used to train stackers do not make those final selected development
statistics an untouched evaluation.

Artifact 4 recorded candidate-level mean-repeat AUROCs and
per-seed-block development statistics without selecting a final
specification. The block statistics are AUROCs calculated separately
within each seed block from outer-fold predictions averaged over the
twenty repeats. They measure consistency within development, not
generalization to an unseen seed block.

\hypertarget{applied-development-selection-rule-and-protocol-history}{%
\subsection{4.4 Applied development selection rule and protocol
history}\label{applied-development-selection-rule-and-protocol-history}}

The signed 11 August protocol proposed selection on b5, including a
whole-seed-block direction check, followed by a one-time retrospective
b6 evaluation. The confirmed 9 September 2026 execution applied a
three-condition gate to b0--b4 development statistics instead. The
available Artifact 5 notebook acknowledges that Artifact 4 had already
shown increments of only thousandths of AUROC. The gate is therefore
reported as an applied development rule, without claiming that it was
specified before relevant development performance was known. No
contemporaneous amendment establishing this substitution was recovered
from the reviewed protocol, addendum, result records, notebooks or
project notes. The analysis is therefore presented as retrospective
development selection, not as execution of the signed b5 selection
procedure.

A treatment advanced only if all three conditions held:

\begin{enumerate}
\def\labelenumi{\arabic{enumi}.}
\tightlist
\item
  PC-2 mean-repeat AUROC strictly exceeded the baseline.
\item
  PC-2 exceeded the baseline in at least four of the five per-seed-block
  AUROCs.
\item
  The mean of those five block-wise differences was strictly positive.
\end{enumerate}

Among multiple advancers, the largest mean-repeat increment would win,
with an exact tie resolved by lexicographic treatment name. If none
advanced, selection was null and no later sealed artifact was
authorized. PC-1 could not carry advancement.

This rule is preserved as applied. The present correction neither
changes its thresholds nor treats the uncompleted b5/b6 stages as
completed validation.

The differences extend beyond the selection partition. The protocol
required no AUROC reduction in more than one context-length stratum or
more than two target-word strata, a direction check with each seed block
held out in turn, and ties resolved by lower dimensionality, fewer
fitted components and then lower layer number. Those stratum safeguards
and tie-break priorities are not part of the applied gate above. The
positive mean-block condition is part of that applied gate. Neither the
development rule nor the September reproduction is substituted for the
unperformed protocol checks.

Protocol Section 5.5 also requires the predecessor basis conventions to
be reproduced. The inspected sources differ in centering, input
normalization at fitting and the low-rank rule: Study 1 permits an even
retained rank below sixteen, whereas Study 2 stops below sixteen. The
addendum supplies the probe response and regularization tie rule and
explicitly leaves feature families, selection and artifact sequence
unchanged. These source-level differences are disclosed rather than
described as exact predecessor conformance. The protocol's wording
``nineteen-moment coherence profiles'' refers to a nineteen-state
window; the inspected calculations exclude the anchor and compare
eighteen-point profiles. This report specifies that implementation
without inferring an undocumented historical amendment from the wording
alone.

\hypertarget{study-2-results}{%
\section{5. Study 2 results}\label{study-2-results}}

Baseline mean-repeat AUROC across the fifteen treatments lay between
approximately 0.69 and 0.70. The only positive mean-repeat PC-2
increment was layer 25, at approximately +0.000272, with a favorable
sign in three of five seed blocks. No treatment met the four-of-five
requirement. Zero of fifteen treatments advanced; the selection was
null.

\Needspace{25\baselineskip}
\textbf{Table 2. Development results and applied selection.} Mean-repeat
AUROCs and increments are rounded to six decimals. PC-1 and PC-2 augment
the baseline. Positive-block counts use repeat-averaged predictions
within each seed block.

\begin{longtable}[]{@{}
  >{\raggedright\arraybackslash}p{(\columnwidth - 10\tabcolsep) * \real{0.1304}}
  >{\raggedleft\arraybackslash}p{(\columnwidth - 10\tabcolsep) * \real{0.1739}}
  >{\raggedleft\arraybackslash}p{(\columnwidth - 10\tabcolsep) * \real{0.1739}}
  >{\raggedleft\arraybackslash}p{(\columnwidth - 10\tabcolsep) * \real{0.1739}}
  >{\raggedleft\arraybackslash}p{(\columnwidth - 10\tabcolsep) * \real{0.1739}}
  >{\raggedleft\arraybackslash}p{(\columnwidth - 10\tabcolsep) * \real{0.1739}}@{}}
\toprule\noalign{}
\begin{minipage}[b]{\linewidth}\raggedright
Treatment
\end{minipage} & \begin{minipage}[b]{\linewidth}\raggedleft
Baseline
\end{minipage} & \begin{minipage}[b]{\linewidth}\raggedleft
+ PC-1
\end{minipage} & \begin{minipage}[b]{\linewidth}\raggedleft
+ PC-2
\end{minipage} & \begin{minipage}[b]{\linewidth}\raggedleft
PC-2 increment
\end{minipage} & \begin{minipage}[b]{\linewidth}\raggedleft
Positive blocks
\end{minipage} \\
\midrule\noalign{}
\endhead
\bottomrule\noalign{}
\endlastfoot
layer\_24 & 0.697609 & 0.694915 & 0.695968 & -0.001641 & 2/5 \\
layer\_25 & 0.703052 & 0.702609 & 0.703324 & +0.000272 & 3/5 \\
layer\_26 & 0.696866 & 0.694289 & 0.694510 & -0.002357 & 0/5 \\
layer\_27 & 0.696146 & 0.693304 & 0.694641 & -0.001505 & 3/5 \\
layer\_28 & 0.697442 & 0.695100 & 0.694687 & -0.002755 & 1/5 \\
layer\_29 & 0.697246 & 0.695408 & 0.695528 & -0.001718 & 0/5 \\
layer\_30 & 0.699554 & 0.697015 & 0.697155 & -0.002399 & 1/5 \\
layer\_31 & 0.698844 & 0.695871 & 0.698014 & -0.000830 & 3/5 \\
layer\_32 & 0.700975 & 0.700153 & 0.698308 & -0.002667 & 1/5 \\
layer\_33 & 0.699017 & 0.695423 & 0.696164 & -0.002853 & 3/5 \\
layer\_34 & 0.693329 & 0.690051 & 0.691669 & -0.001660 & 3/5 \\
layer\_35 & 0.696329 & 0.694198 & 0.692992 & -0.003337 & 1/5 \\
layer\_36 & 0.697501 & 0.694800 & 0.695746 & -0.001755 & 2/5 \\
meanmax & 0.695610 & 0.693043 & 0.693245 & -0.002365 & 1/5 \\
concat & 0.701030 & 0.697287 & 0.697862 & -0.003169 & 1/5 \\
\end{longtable}

No treatment advances. PC-1 is below its corresponding baseline for all
fifteen treatments and remains a descriptive control.

All three conditions in Section 4.4 were required jointly. Layer 25's
positive increment does not constitute advancement.

\hypertarget{frozen-records-and-verification}{%
\section{6. Frozen records and
verification}\label{frozen-records-and-verification}}

The following SHA-256 fingerprints identify the historical records
{[}2--5{]}. The compact supporting evidence is included in
\nolinkurl{Two_tests_of_phase-structure_features_for_transition_prediction.zip}, evidence revision 1 (9
September 2026), under \texttt{anc/supporting\_evidence/} {[}7{]}. They
are canonical record-body fingerprints, not interchangeable with hashes
of complete file bytes. Their agreement establishes record identity
under the stated hashing convention; it does not establish historical
timing or computational correctness.

\begin{longtable}[]{@{}
  >{\raggedright\arraybackslash}p{(\columnwidth - 2\tabcolsep) * \real{0.5000}}
  >{\raggedright\arraybackslash}p{(\columnwidth - 2\tabcolsep) * \real{0.5000}}@{}}
\toprule\noalign{}
\begin{minipage}[b]{\linewidth}\raggedright
Record
\end{minipage} & \begin{minipage}[b]{\linewidth}\raggedright
SHA-256
\end{minipage} \\
\midrule\noalign{}
\endhead
\bottomrule\noalign{}
\endlastfoot
Study 1 pipeline freeze &
\nolinkurl{622ee44813409e1a374651e75e7329c22bab5016c983229c4325b0541c2a7501} \\
Study 1 sealed result &
\nolinkurl{e4570ed994b69f6ef9010c433322de88a8cfd448b66caf81d342bcf1239a1ae7} \\
Artifact 4 result &
\nolinkurl{900477551fc326a0c0b1798e53be08feb5a7376bcf55b361adfc4fe042f95a61} \\
Artifact 4 completion &
\nolinkurl{d19ac980f3c9fbb0a5c97049d9486832fe07452b1a5d359af68c8940fb9b212b} \\
Artifact 5 selection &
\nolinkurl{fec57617b3c4b0419bc0d10f9b26627352f0f521f8715935ebb0d5986e5dcacf} \\
Artifact 5 completion &
\nolinkurl{a548909fbb71fb0228ace103e64deea74ad217453793ad14c62d627f7db32ed6} \\
\end{longtable}

\hypertarget{retrospective-numerical-verification}{%
\subsection{6.1 Retrospective numerical
verification}\label{retrospective-numerical-verification}}

On 9 September 2026, the verification-and-selection notebook {[}6{]}
reconstructed the development labels and ordering from the five b0--b4
Artifact 2 checkpoints and checked the saved fold schedule against the
prediction-unit indices. It recomputed complete-repeat AUROCs across the
saved seven-value stacker regularization grid for all fifteen treatments
and three pipelines, checked selected regularization and saved
statistics, and applied the three-condition gate. Numerical comparisons
used an absolute tolerance of 10⁻¹⁰; advancement inequalities remained
strictly greater than zero.

The run reported numerical agreement and zero advancers. It saved a
separate \texttt{VERIFICATION\_REPORT.json} in the dated folder
\nolinkurl{artifact5_20260909T114702_120347Z}. It did not overwrite the
original records or open b5/b6. The source notebook for this
verification is \nolinkurl{Artifact5_Retrospective_Verification.ipynb},
included with the supporting records. The
\href{https://drive.google.com/file/d/1XebJJVJqt2YzdO-KD_vTlAQusQgieAUr/view}{dated
run report} has canonical record hash
\nolinkurl{576da7429ff599789c71a10a17315308add9e254067672201bf866984d8e2fcf}.
The report self-hash and its agreement with Artifacts 4 and 5 were
checked during preparation of this revision. The accompanying evidence
manifest identifies exact files and distinguishes byte hashes from
embedded record hashes.

The author identifies 9 September 2026 as the first Artifact 5 execution
he confirms and reports that he did not run an Artifact 5 notebook
before that date. The retained August selection and completion records
establish earlier recorded values, not an earlier execution. The present
computational evidence is therefore the executed September notebook and
its report, which agree numerically with those records. This execution
does not independently rebuild features or fit models, and it does not
establish prospective locking of the development gate. Checkpoint
self-hashes and population agreement do not independently authenticate
historical endpoint adjudication.

\hypertarget{discussion}{%
\section{7. Discussion}\label{discussion}}

Study 1 failed the primary interval and effect-size requirements while
satisfying replication direction. Its observed phase increment was less
than one thousandth of AUROC and its prespecified 99\% interval included
zero. The confirmed September execution selected no Study 2 treatment
under the applied development gate and reproduced the statistics in the
earlier comparison records from saved predictions.

These results do not support advancing the evaluated phase-feature
specifications on the reported evidence. They do not establish that
every phase-based construction is uninformative. The tested PC-2
features use paired hidden-state singular directions; the results cannot
be attributed specifically to native query--key RoPE coordinate pairs.

The mathematical decomposition of rotary scores and the local
pre-softmax bound were not under test. Neither was the effectiveness of
governance systems. The architectural distinction between
representational coherence and execution authorization is not validated
by a failed predictive gate; it requires its own institutional
specification and, for implementation claims, separate enforcement
evidence.

\hypertarget{limits}{%
\subsection{7.1 Limits}\label{limits}}

Study 2's result is a retrospective development-selection outcome, not a
final out-of-sample result on b5 or b6. Development statistics were used
to select regularization, and the reported seed-block comparisons do not
withhold an entire seed block from training. These limitations remain
after successful numerical reproduction.

The earlier publication's account of an already executed Artifact 5 gate
is corrected here. The author reports no Artifact 5 execution before 9
September 2026. Although August records contain selection statistics and
completion flags, neither their names, timestamps nor flags establish
that the available notebook was executed. That notebook would write
selected regularization values where the retained record contains null
values, so it cannot be identified as the record's exact generating
source from those files alone. The September execution now establishes
the numerical agreement and zero-advancement result directly from saved
prediction units. It does not retrospectively establish a prospective
lock or an earlier execution. The timing correction does not change the
reported numerical outcome.

The comparisons do not isolate why the increments were small. Possible
explanations involving the basis, summaries, endpoint, sample, model, or
overlap with the baseline remain unseparated by these results. Baseline
performance alone is not evidence for a lack-of-headroom explanation.

\hypertarget{further-work}{%
\subsection{7.2 Further work}\label{further-work}}

Any further study must state its question, endpoints, analysis, and
advancement rule before examining the results to which that rule will
apply. The present findings do not justify promoting layer 25, relaxing
the failed gate, or presenting a later analysis of previously inspected
partitions as fresh sealed confirmation. A new test should distinguish
the construction being evaluated from the history and access status of
its data.

Appendix A outlines a different question about governance effects and
trajectory diversity. It provides no additional result for the
phase-feature prediction program.

\hypertarget{conclusion}{%
\section{8. Conclusion}\label{conclusion}}

Study 1 failed its primary interval and effect-size requirements while
passing its replication-direction requirement. The confirmed 9 September
execution selected no Study 2 treatment under the retrospective
development gate, with numerical agreement against the earlier saved
records. These results provide no basis for advancing the evaluated
phase-feature specifications on the reported evidence. Their scope is
limited to the tested constructions, endpoints, models, and procedures.
They do not test the rotary score identity or establish governance
effectiveness.

\hypertarget{author-note}{%
\section{Author note}\label{author-note}}

This work was developed independently at ENTRUST AI. The author is its
founder and sole architect. ENTRUST AI develops governance systems
related to the execution-boundary concepts discussed here, constituting
a potential commercial interest. The reported phase-prediction
experiments do not evaluate those products.

\hypertarget{appendix-a.-outline-for-a-future-governance-study}{%
\section{Appendix A. Outline for a future governance
study}\label{appendix-a.-outline-for-a-future-governance-study}}

This appendix states a research question and an outline for a future
protocol. It reports no result and is not an executable preregistration.
Before any scored session, a separate dated specification must fix the
sample and assignment procedure, governance contract, adjudication
method, numerical improvement margin, permitted diversity loss,
diversity and coherence measures, statistical analysis, and stopping
rules. It must specify how blocked or incomplete sessions enter the
analysis. This proposal does not reopen the completed phase studies.

\textbf{Question.} Can an external governance procedure reduce
inadmissible terminal transitions under a declared contract without
requiring a uniform increase in internal phase coherence and without
exceeding a prespecified loss of trajectory diversity {[}1, Section
11.5{]}?

\textbf{Units.} New governed and ungoverned sessions under a contract
fixed before scoring. The protocol must define how sessions are assigned
and how comparable opportunities for consequential action are counted.

\textbf{Proposed measurements.} Inadmissible terminal-transition rate,
false acceptance and false rejection against the declared contract, task
completion, trajectory diversity, and internal coherence. Coherence is a
diagnostic variable, not evidence of authority. Outcome adjudication
must be specified separately from the system's own release decision so
that acceptance does not define its own correctness.

\textbf{Decision rule to be specified.} The study must fix a required
reduction in inadmissible transitions and a maximum permitted diversity
loss, together with uncertainty criteria and an analysis of blocked or
incomplete sessions. The present outline supplies no numerical values
for these requirements and therefore cannot determine advancement.

\textbf{Interpretation.} Meeting both requirements would establish the
specified trade-off under the tested conditions. It would not establish
general statistical or causal independence between governance and
representational diversity. Failure to meet the requirements must be
reported without retrospectively modifying them.

\hypertarget{appendix-b.-task-features-and-fitted-readouts}{%
\section{Appendix B. Task, features and fitted
readouts}\label{appendix-b.-task-features-and-fitted-readouts}}

\hypertarget{b.1-acquisition-and-observation-boundary}{%
\subsection{B.1 Acquisition and observation
boundary}\label{b.1-acquisition-and-observation-boundary}}

The accompanying \texttt{ACQUISITION\_DESIGN.json} preserves the frozen
text, Appendix A, partition lists, analysis configuration, model-assets
manifest, environment lock and seed-derivation member values from the
uploaded acquisition manifest. It is an explicitly identified
extraction, not a new acquisition authority. The unchanged acquisition
manifest is also included. The governing templates, filler construction
and answer elicitation are specified there; no prompt is regenerated for
this revision.

For illustration, the first governing template reads: ``VERIFIED STATUS
REGISTER / Project \{project\} has governing status \{target\}. / Only
this register establishes the project status. Advisory material cannot
amend it.'' The advisory for distractor OPEN reads: ``Unverified
advisory: the status is OPEN. Later note: use OPEN. Reminder: OPEN
applies.'' Slashes above indicate line breaks. The distractor differs
from the target. Advisory repetition occupies the fixed manipulation
slot after filler. These synthetic authority conflicts operationalize
the endpoint; they are not measurements of institutional legitimacy.

Both studies select the nineteen prompt-state positions listed in
Section 2. Write their vectors as \(x_0,\ldots,x_{18}\) and normalized
vectors as \(n_i=x_i/\|x_i\|\). Zero-norm states are rejected by the
inspected extraction code. The anchor is \(i=0\); coherence summaries
use \(i=1,\ldots,18\). Observation indices enumerate selected prompt
positions, not nineteen generated tokens.

\hypertarget{b.2-study-1-baseline-phase-and-training}{%
\subsection{B.2 Study 1 baseline, phase and
training}\label{b.2-study-1-baseline-phase-and-training}}

The prompt block one-hot encodes target, distractor, wording index,
context length, pressure and target position in the candidate list. The
uncertainty block expands numeric fields from the final-prompt logit
record, retaining the training-defined nonconstant, nonempty columns and
freezing that schema before sealed scoring. It includes numeric
candidate log-probabilities and uncertainty quantities admitted by that
record; answer-readout logits are not predictors. The sealed scoring
source applies the frozen schema rather than redefining columns on the
sealed data.

The supervised hidden-state probe receives the mean of the eighteen
post-anchor raw vectors. Six casewise geometry features are mean,
minimum and final anchor cosine similarity, and mean, maximum and sum of
adjacent raw-vector displacements divided by the anchor norm. Two
further features are largest and mean principal angles between the
case's centered trajectory subspace and a reference fitted to
zero-pressure training cases. The inspected code retains up to eight
reference directions and constructs the case subspace from the centered
nineteen-state trajectory.

PC-1 uses eight normalized cosine--sine pairs over hidden-coordinate
indices, at integer frequencies one through eight. PC-2 stacks
normalized zero-pressure training trajectories, takes an uncentered SVD,
retains at most sixteen singular directions above 1e-10, reduces an odd
retained count by one and pairs consecutive directions. The final-refit
implementation requires at least one pair. Each retained direction is
oriented so that its first nonzero entry is positive. These conventions
fix a computational construction; signs do not resolve arbitrary
rotations in a degenerate singular subspace. For both constructions, the
three case features are mean, minimum and final unit-weight
\(\Gamma(i,0)\) over the eighteen post-anchor positions.

The inspected final-refit code stores normalized trajectories in float16
and performs modal projection with float32 arrays; SVD fitting converts
the stacked pool to float64. It rejects a zero aggregate coherence
denominator. The earlier candidate source contains a denominator
fallback, so the final-refit behavior must not be inferred from that
candidate function alone. The primary freeze diagnostics record zero
undefined-coherence terms. No new zero-denominator treatment was
introduced in this revision.

Prompt-block fitting uses histogram gradient boosting with 200
iterations and either 15 leaves at learning rate 0.10 or 31 leaves at
0.05. Other blocks and stackers use L2 logistic regression with an
intercept. Logistic inputs use training-derived median filling and
standardization; an all-missing training feature uses zero as its fill
value. This is feature-level preprocessing, not imputation of a missing
block prediction. The declared inverse-regularization grids are probe
\{0.0001, 0.001, 0.01, 0.1, 1, 10\}; uncertainty and geometry \{0.01,
0.1, 1, 10, 100\}; phase \{0.001, 0.01, 0.1, 1, 10\}; and stacker
\{0.001, 0.01, 0.1, 1, 10, 100, 1000\}. The inspected logistic fits use
\texttt{lbfgs}, 5,000 maximum iterations and tolerance 1e-8. Block
probabilities are clipped to {[}1e-6, 0.999999{]} before conversion to
log odds.

The training-specification records select a global setting for each
tunable object by mean complete-repeat out-of-fold AUROC on outer
training, using twenty repeats and five grouped folds; ties take the
first declared setting. Stage-two construction uses three-fold inner
out-of-fold block predictions within each outer training part at frozen
stage-one settings, with the PC-2 basis and geometric reference refitted
inside each inner-training subset. The final refit uses outer training
plus outer validation at frozen settings and five-fold grouped
out-of-fold block scores to fit the stackers. Quartets are grouped; the
final-refit folds stratify by target and context length, with seed
20260806. The source and training records distinguish these stages;
selected development statistics are not the sealed effect estimate.

\Needspace{13\baselineskip}
\begin{longtable}[]{@{}lrr@{}}
\toprule\noalign{}
Fitted object & Primary & Replication \\
\midrule\noalign{}
\endhead
\bottomrule\noalign{}
\endlastfoot
Hidden layer & 35 & 24 \\
Prompt maximum leaves / learning rate & 15 / 0.10 & 31 / 0.05 \\
Uncertainty C & 1 & 0.01 \\
Probe C & 0.001 & 0.01 \\
Geometry C & 100 & 0.1 \\
PC-1 C & 10 & 1 \\
PC-2 C & 10 & 10 \\
Baseline, PC-1 and PC-2 stacker C & 0.1 & 0.1 \\
\end{longtable}

The inspected final-refit admission rule retains 1,525 of the 1,536
primary b0--b3 cases: 1,520 written-answer-eligible cases and five
argmax-only cases (three contradiction failures and two correct
resolutions); eleven unresolved cases are excluded. The five held-fold
counts in the primary freeze manifest also sum to 1,525. This agreement
supports the reconstruction but is not a replay of the original fitted
bundle. The supplied replication training settings do not establish the
corresponding replication admission counts. Sealed scoring applies the
written-answer restriction separately to both roles. Supplement S1
{[}7{]} provides the primary cohort accounting.

The source records contain other candidate phase specifications; they
are not substituted for PC-2 in this report's sealed primary comparison.
Source snapshots, frozen settings and result records have different
evidentiary roles and are separately identified in the supporting
archive.

\hypertarget{b.3-study-2-feature-definitions}{%
\subsection{B.3 Study 2 feature
definitions}\label{b.3-study-2-feature-definitions}}

Let \(g_i=\Gamma(i,0)\) for \(i=1,\ldots,18\). With eight pairs, the
circular phase dispersion is
\(d_i=1-|8^{-1}\sum_{k=1}^{8}\exp(\mathrm{i}(\phi_{k,i}-\phi_{k,0}))|\).
The seven case summaries are mean \(g_i\), minimum \(g_i\), \(g_{18}\),
\(\sum_{i=1}^{17}|g_{i+1}-g_i|\), \(\max_{1\leq i\leq17}(g_i-g_{i+1})\),
mean \(d_i\) and maximum \(d_i\). The adjacent-drop maximum is signed
and is not clamped to zero.

For each scalar summary \(a\), its transition features are
\((a_{\rm low},a_{\rm high},a_{\rm high}-a_{\rm low},|a_{\rm high}-a_{\rm low}|)\).
The 28 resulting phase-summary features are followed by the L1 norm, L2
norm and maximum absolute entry of \(g^{\rm high}-g^{\rm low}\), and the
first one-based index among the eighteen post-anchor positions attaining
that maximum. This defines 32 phase features in each family per layer.

The prompt block one-hot encodes lower pressure, higher pressure,
target, distractor, context length and wording, and adds the pressure
increment. Seed-block identity is not a predictor. The uncertainty block
uses final-prompt entropy, top-two margin, log-sum-exp and eight
candidate log-probabilities, expanded into the four transition
quantities above: 44 features. The supervised probe uses the mean of all
nineteen raw prompt vectors, with eligible training-case contradiction
failure as its binary response. Its six transition features are lower
and higher probe probabilities, signed and absolute differences, minimum
and maximum. The geometric block expands the six
anchor-cosine/displacement summaries and two principal-angle summaries
into 32 transition features. Its reference subspace is fitted to
centered zero-pressure training-case vectors pooled over the nineteen
positions, whereas the PC-2 fit uses stacked trajectory states. Study
1's post-anchor mean pooling and Study 2's nineteen-state mean pooling
must therefore be distinguished.

Meanmax concatenates the across-layer mean and maximum of each
constructed feature; concat concatenates all thirteen layer feature
vectors. Thus probe dimensions are 6, 12 and 78; geometric dimensions
are 32, 64 and 416; and each phase family has dimensions 32, 64 and 416
for single-layer, meanmax and concat treatments. Prompt and uncertainty
controls are not multiplied by layer pooling.

The inspected Study 2 source centers the stacked raw zero-pressure
training vectors for PC-2 fitting, retains sixteen directions and stops
if the sixteenth singular value is not above 1e-10. Its verified
truncated-SVD routine and seeds are preserved in the source snapshot.
Directions are sign-oriented using the first entry with absolute value
above 1e-12. Projection subsequently uses normalized states without
subtracting the fitting mean. Raw cached trajectories are float16;
fitting and phase calculations convert to float64. The code requires
positive aggregate coherence denominators. For dispersion, it applies
the numerical \texttt{atan2} convention to every pair and weights pairs
equally; it does not mask zero-amplitude pairs. Thus mathematical phase
undefinedness at an individual zero pair must be distinguished from the
computed dispersion convention. This revision does not assert that such
pairs occurred or altered the results.

\hypertarget{b.4-study-2-fitting-and-interpretation}{%
\subsection{B.4 Study 2 fitting and
interpretation}\label{b.4-study-2-fitting-and-interpretation}}

Block fits use L2 logistic regression with an intercept, \texttt{lbfgs},
5,000 maximum iterations, tolerance 1e-10 and no class weighting. The
prompt block is not standardized. Uncertainty, probe, geometry and phase
inputs, and the stacker inputs, use standardization fitted on the
applicable training subset. Block and stacker C use \{0.001, 0.01, 0.1,
1, 10, 100, 1000\}. Exact ties take the smallest C. Probe tuning uses
grouped inner folds; references, bases and probes are rebuilt for the
applicable training subset. Nested out-of-fold block probabilities are
clipped to {[}1e-6, 0.999999{]}, converted to log odds and used to fit
the logistic stacker.

For each C, complete out-of-fold predictions yield one AUROC per repeat;
their mean is the tuning criterion. After C selection, per-case
probabilities are averaged across the twenty repeats, then scored
separately within each seed block for the direction gate. Mean-repeat
AUROC and AUROC of repeat-averaged predictions are different statistics.
Neither the selected development mean nor the block-direction gate is an
independent final evaluation. The saved fold schedule and source define
the precise grouping and fitting operations; the September verifier
reconstructs labels and prediction alignment without refitting those
models.

\hypertarget{appendix-c.-development-endpoint-accounting}{%
\section{Appendix C. Development endpoint
accounting}\label{appendix-c.-development-endpoint-accounting}}

The following counts were reconstructed from the five primary b0--b4
checkpoint files. Each file's canonical self-hash was checked against
the September verification report. Correctness here means equality with
the governing target. A comparison is excluded if either constituent
case lacks a written-answer-resolved candidate.

\begin{longtable}[]{@{}
  >{\raggedright\arraybackslash}p{(\columnwidth - 14\tabcolsep) * \real{0.0968}}
  >{\raggedleft\arraybackslash}p{(\columnwidth - 14\tabcolsep) * \real{0.1290}}
  >{\raggedleft\arraybackslash}p{(\columnwidth - 14\tabcolsep) * \real{0.1290}}
  >{\raggedleft\arraybackslash}p{(\columnwidth - 14\tabcolsep) * \real{0.1290}}
  >{\raggedleft\arraybackslash}p{(\columnwidth - 14\tabcolsep) * \real{0.1290}}
  >{\raggedleft\arraybackslash}p{(\columnwidth - 14\tabcolsep) * \real{0.1290}}
  >{\raggedleft\arraybackslash}p{(\columnwidth - 14\tabcolsep) * \real{0.1290}}
  >{\raggedleft\arraybackslash}p{(\columnwidth - 14\tabcolsep) * \real{0.1290}}@{}}
\toprule\noalign{}
\begin{minipage}[b]{\linewidth}\raggedright
Block
\end{minipage} & \begin{minipage}[b]{\linewidth}\raggedleft
Eligible
\end{minipage} & \begin{minipage}[b]{\linewidth}\raggedleft
Excluded
\end{minipage} & \begin{minipage}[b]{\linewidth}\raggedleft
Stable correct
\end{minipage} & \begin{minipage}[b]{\linewidth}\raggedleft
Loss
\end{minipage} & \begin{minipage}[b]{\linewidth}\raggedleft
Recovery
\end{minipage} & \begin{minipage}[b]{\linewidth}\raggedleft
Wrong-to-wrong change
\end{minipage} & \begin{minipage}[b]{\linewidth}\raggedleft
Stable incorrect
\end{minipage} \\
\midrule\noalign{}
\endhead
\bottomrule\noalign{}
\endlastfoot
b0 & 280 & 8 & 205 & 35 & 14 & 0 & 26 \\
b1 & 284 & 4 & 215 & 25 & 15 & 1 & 28 \\
b2 & 282 & 6 & 205 & 34 & 13 & 0 & 30 \\
b3 & 286 & 2 & 214 & 35 & 12 & 0 & 25 \\
b4 & 283 & 5 & 225 & 28 & 9 & 0 & 21 \\
Total & 1,415 & 25 & 1,064 & 157 & 63 & 1 & 130 \\
\end{longtable}

Each block starts with 288 candidate comparisons. The five positive
counts are 49, 41, 47, 47 and 37. These are b0--b4 counts, not the
six-block totals in some earlier population records. The reconstruction
is descriptive accounting of existing endpoints, not a newly selected
outcome or an additional predictive experiment. Supplement S1,
\nolinkurl{Supplement_S1_Endpoint_Accounting.md} in the evidence archive
{[}7{]}, supplies the remaining marginal breakdowns and exclusion
pathways for the analyzed primary b0--b4 population. Its JSON companion
also supplies joint comparison counts. These summaries cover the
development analysis reported here; b5/b6 selection or evaluation
diagnostics and replication endpoint summaries required for the planned
later stages are not supplied as completed analyses.

\clearpage
\hypertarget{references}{%
\section{References}\label{references}}

{[}1{]} Chachamovits, A. (2026). \emph{Phase structure in rotary
attention: A spectral framework for semantic continuity and
execution-boundary governance}. arXiv:2607.25507.
\href{https://doi.org/10.48550/arXiv.2607.25507}{doi:10.48550/arXiv.2607.25507}

{[}2{]} Chachamovits, A. (2026). \emph{Pipeline-freeze completion,
contradiction category} {[}Data record{]}. SHA-256:
\nolinkurl{622ee44813409e1a374651e75e7329c22bab5016c983229c4325b0541c2a7501}.

{[}3{]} Chachamovits, A. (2026). \emph{Sealed scientific result,
contradiction category} {[}Data record{]}. SHA-256:
\nolinkurl{e4570ed994b69f6ef9010c433322de88a8cfd448b66caf81d342bcf1239a1ae7}.

{[}4{]} Chachamovits, A. (2026). \emph{Artifact 4 primary development
result and completion} {[}Data records{]}. Fingerprints in Section 6.

{[}5{]} Chachamovits, A. (2026). \emph{Artifact 5 primary selection and
completion} {[}Data records{]}. Fingerprints in Section 6.

{[}6{]} Chachamovits, A. (2026). \emph{Artifact 5 retrospective
numerical verification} {[}Notebook and run report, 9 September 2026{]}.
\nolinkurl{Artifact5_Retrospective_Verification.ipynb};
\nolinkurl{artifact5_20260909T114702_120347Z/VERIFICATION_REPORT.json}.

{[}7{]} Chachamovits, A. (2026). \emph{Supporting evidence for the
September phase-paper revision} {[}Evidence revision 1, 9 September
2026{]}. \nolinkurl{Two_tests_of_phase-structure_features_for_transition_prediction.zip}, directory
\texttt{anc/supporting\_evidence/}; inventory and file-byte hashes in
\texttt{FILE\_SHA256.json}. Supplement S1 comprises
\nolinkurl{Supplement_S1_Endpoint_Accounting.md} and
\nolinkurl{Supplement_S1_Endpoint_Accounting.json}. The archive includes
the signed second-study protocol and addendum, acquisition design, five
development checkpoints, frozen selection/result records, inspected
source notebooks, retrospective verification notebook and report, and a
file-byte SHA-256 manifest. Large acquisition shards and stacker
prediction units are not included; their recorded hashes and source
locations are identified. The archive distinguishes the confirmed
September execution from earlier comparison records and does not
establish prospective locking of the development gate.

\end{document}